\documentclass[10pt,twocolumn,letterpaper]{article}

\usepackage[pagenumbers]{cvpr} %

\usepackage[dvipsnames]{xcolor}

\usepackage[utf8]{inputenc}
\usepackage{tabularx}
\usepackage{graphicx}
\usepackage{booktabs}
\usepackage{colortbl}
\usepackage{microtype}
\usepackage[accsupp]{axessibility}
\usepackage{pgfplots}
\pgfplotsset{compat=1.18}
\usepackage[graphicx]{realboxes}
\usepackage{bm}
\usepackage{nicefrac}
\usepackage{multirow}
\usepackage[number-mode=text]{siunitx}
\usepackage[scaled=.97]{newtxtt}
\usepackage{marvosym}
\usepackage{placeins}
\usepackage{xr}
\usepackage{xpatch}
\usepackage{fancyhdr}
\usepackage{setspace}

\usepackage{pifont}
\newcommand{\cmark}{\text{\ding{51}}}

\newcommand*{\tran}{^{\mkern-1.5mu\mathsf{T}}}

\newcommand*{\inparagraph}[1]{\smallskip\noindent\textbf{#1}\hspace{0.5em}}

\usepackage{xspace}
\newcommand{\adapter}{Adapter+\xspace}

\newcolumntype{+}{>{\global\let\currentrowstyle\relax}}
\newcolumntype{-}{>{\currentrowstyle}}
\newcommand{\rowstyle}[1]{\gdef\currentrowstyle{#1}\leavevmode#1\ignorespaces}
\newcommand{\tblgray}{gray!75}
\newcommand{\grayrow}{\rowstyle{\color{\tblgray}}}

\usepackage{trimclip}
\makeatletter
\DeclareRobustCommand{\circbullet}{\mathbin{\vphantom{\circ}\text{\circbullet@}}}
\newcommand{\circbullet@}{%
  \check@mathfonts
  \m@th\ooalign{%
    \clipbox{0 0 0 {\dimexpr\height-\fontdimen22\textfont2}}{$\bullet$}\cr
    $\circ$\cr
  }%
}
\makeatother

\DeclareSIUnit{\pp}{\textup{pp}}

\definecolor{cvprblue}{rgb}{0.21,0.49,0.74}
\usepackage[breaklinks,colorlinks,citecolor=cvprblue]{hyperref}

\def\confName{CVPR}
\def\confYear{2024}

\title{Adapters Strike Back}

\author{Jan-Martin O. Steitz$^1$ \hspace{1cm} Stefan Roth$^{1,2}$\\
$\ ^1$Department of Computer Science, TU Darmstadt \hspace{1cm} $\ ^2$ hessian.AI}

\definecolor{tabhighlight}{HTML}{e5e5e5}
\newcommand{\tableCellHeight}{1}
\newcommand{\tabstyle}[1]{
  \setlength{\tabcolsep}{#1}
  \renewcommand{\arraystretch}{\tableCellHeight}
  \centering
  \small
}

\begin{document}
\maketitle
\thispagestyle{fancy}
\definecolor{lora}{HTML}{00BFFF}
\definecolor{fact}{HTML}{D4CA3A}
\definecolor{vpt}{HTML}{FF6DAE}
\definecolor{ssf}{HTML}{00B78D}
\definecolor{cons}{HTML}{BEA9FF}
\definecolor{spt}{HTML}{FF6765}
\definecolor{adp}{HTML}{63DF75}
\definecolor{arxiv}{HTML}{F0F0F0}

\definecolor{gray1}{gray}{0.9}
\definecolor{gray2}{gray}{0.72}

\renewcommand*{\thefootnote}{\fnsymbol{footnote}}
\begin{abstract}
    Adapters provide an efficient and lightweight mechanism for adapting trained transformer models to a variety of different tasks.
    However, they have often been found to be outperformed by other adaptation mechanisms, including low-rank adaptation. 
    In this paper, we provide an in-depth study of adapters, their internal structure, as well as various implementation choices. 
    We uncover pitfalls for using adapters and suggest a concrete, improved adapter architecture, called \emph{\adapter}, that not only outperforms previous adapter implementations but surpasses a number of other, more complex adaptation mechanisms in several challenging settings. 
    Despite this, our suggested adapter is highly robust and, unlike previous work, requires little to no manual intervention when addressing a novel scenario. 
    \adapter reaches state-of-the-art average accuracy on the VTAB benchmark, even without a per-task hyperparameter optimization.\footnote[2]{Code is available at \url{https://github.com/visinf/adapter_plus}.}
\end{abstract}

\section{Introduction}
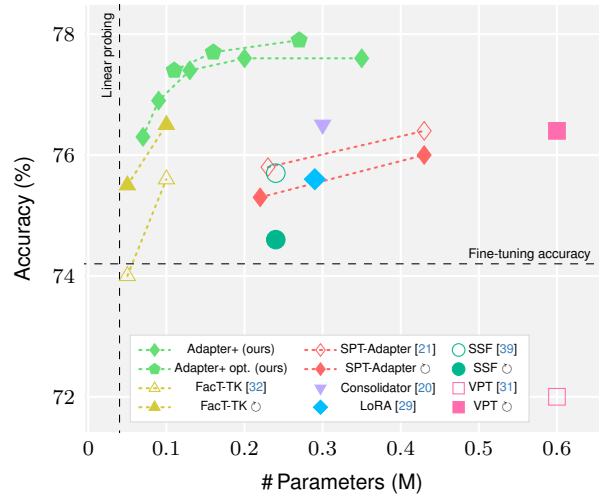
\begin{figure}
    \centering
    \def\marksize{3.5pt}
\begin{tikzpicture}
    \begin{axis}[
        axis background/.style={fill=gray!10},
        grid,
        every major grid/.style = {style={white, thick}},
        tick label style={major tick length=0pt},
        axis line style={white},
        legend image post style={scale=0.68}, %
        legend style={
            font=\sffamily\scriptsize,
            nodes={scale=0.68, transform shape}, %
            at={(0.088,0.126)},
            anchor=west,
            draw=gray!30,
            /tikz/every even column/.append style={column sep=0.17cm}
        },
        legend columns=4,
        transpose legend,
        xlabel={\#\,Parameters (M)},
        ylabel={Accuracy (\%)},
        font=\sffamily\footnotesize,
    ]%
        \addplot[ %
            adp,
            dotted,
            line cap=round,
            thick,
            mark=diamond*,
            mark size=\marksize,
            mark options={solid, thin},
        ] coordinates {(0.07,76.3)(0.09,76.9)(0.13,77.4)(0.20,77.6)(0.35,77.6)};
        \addplot[ %
            adp,
            dotted,
            line cap=round,
            thick,
            mark=pentagon*,
            mark size=0.9*\marksize,
            mark options={solid, thin},
        ] coordinates {(0.11,77.4)(0.16,77.7)(0.27,77.9)};
        
         \addplot[ %
            fact,
            dotted,
            line cap=round,
            thick,
            mark=triangle,
            mark size=\marksize,
            mark options={solid, thin},
        ] coordinates {(0.05,74.0)(0.1,75.6)};
        \addplot[ %
            fact,
            dotted,
            line cap=round,
            thick,
            mark=triangle*,
            mark size=\marksize,
            mark options={solid, thin},
        ] coordinates {(0.05,75.5)(0.1,76.5)};
        
        \addplot[ %
            spt,
            dotted,
            line cap=round,
            thick,
            mark=diamond,
            mark size=\marksize,
            mark options={solid, thin},
        ] coordinates {(0.23,75.8)(0.43,76.4)};
        \addplot[ %
            spt,
            dotted,
            line cap=round,
            thick,
            mark=diamond*,
            mark size=\marksize,
            mark options={solid, thin},
        ] coordinates {(0.22,75.3)(0.43,76.0)};
        
        \addplot[ %
            cons,
            only marks,
            mark=triangle*,
            mark options={rotate=180},
            mark size=1.0*\marksize
        ] coordinates {(0.3,76.5)};
        \addplot[ %
            lora,
            only marks,
            mark=square*,
            mark options={rotate=45},
            mark size=0.8*\marksize
        ] coordinates {(0.29,75.6)};
        
        \addplot[ %
            ssf,
            only marks,
            mark=o,
            mark size=\marksize
        ] coordinates {(0.24,75.7)};
        \addplot[ %
            ssf,
            only marks,
            mark size=\marksize
        ] coordinates {(0.24,74.6)};

        \addplot[ %
            vpt,
            only marks,
            mark=square,
            mark size=0.9*\marksize
        ] coordinates {(0.6,72.0)};
        \addplot[ %
            vpt,
            only marks,
            mark=square*,
            mark size=0.9*\marksize
        ] coordinates {(0.6,76.4)};
        
        \draw[color=black, dashed] 
                    (axis cs:0.04, 70) -- (axis cs:0.04, 80);
        \draw[color=black, dashed] 
                    (axis cs:-0.5, 74.2) -- (axis cs:1.0, 74.2);
        \node at (0.565,74.35) {\tiny Fine-tuning accuracy};
        \node at (0.0255,77.65) {\tiny\rotatebox{90}{Linear probing}};

        \legend{\adapter (ours), \adapter opt. (ours), FacT-TK \cite{Jie:2022:FFT}, FacT-TK $\circlearrowright$, SPT-Adapter \cite{He:2023:SAV}, SPT-Adapter $\circlearrowright$, Consolidator \cite{Hao:2023:CMA}, LoRA \cite{Hu:2022:LLR}, SSF \cite{Lian:2022:SSY}, SSF $\circlearrowright$, VPT \cite{Jia:2022:VPT}, VPT $\circlearrowright$}
    \end{axis}
\end{tikzpicture}
    \vspace{-0.8cm}
    \caption{\textbf{Parameter-accuracy characteristics of adaptation methods on the VTAB \cite{Zhai:2020:LSR} \emph{test sets}.} We report original results and re-evaluations ($\circlearrowright$) after a complete training schedule with suitable data normalization. Our \adapter has clearly the best parameter-accuracy trade-off. The vertical, dashed line shows the possible minimal number of tunable parameters when only the classifiers are trained, \ie, using linear probing (61\% accuracy).}
    \label{fig:vtab_params_accuracy}
    \vspace{-0.5em}
\end{figure}
Transfer learning from an off-the-shelf model, pre-trained on a large dataset like ImageNet \cite{Russakovsky:2015:ILS} to a downstream task by fully fine-tuning the model's parameters is a common paradigm.
A typical CNN architecture, like a ResNet \cite{He:2016:DRL}, has several tens of millions of parameters.
However, since the introduction of transformers \cite{Vaswani:2017:AAY} into the realm of computer vision \cite{Dosovitskiy:2021:IWW,Carion:2020:EEO,Caron:2021:EPS,Ranftl:2021:VTD,Xie:2021:SSE,Radford:2021:LTV}, model sizes have grown exponentially from around a hundred million parameters for a vision transformer (ViT) \cite{Dosovitskiy:2021:IWW} to more than a billion parameters \cite{Dehghani:2023:SVT,Oquab:2023:DLR}.
This leads to huge storage requirements when fine-tuning on multiple downstream tasks because a complete set of the model's parameters needs to be saved per task.
Additionally, large models require correspondingly large datasets \cite[\eg,][]{Schuhmann:2022:LOL} to be trained to their full potential, yet tend to overfit easily if the target dataset in transfer learning is too small.
One solution is linear probing \cite{Donahue:2014:DDC}, where only the linear classifier is trained, but this usually yields inferior results compared to full fine-tuning.

As a consequence, there is a growing interest in parameter-efficient tuning methods. The main idea is to freeze the parameters of the pre-trained model and add a comparatively small amount of parameters to the model, which are then tuned together with the classifier to adapt the model to the downstream task at hand.
Representative methods with different underlying concepts include VPT \cite{Jia:2022:VPT}, which prepends the sequence of image tokens in the attention with trainable tokens to learn a prompt tuning, LoRA \cite{Hu:2022:LLR}, where the attention weights are updated with learnable low-rank decomposition matrices, and Adapters \cite{Houlsby:2019:PET}, which are small bottleneck modules that are added to every transformer layer of the network.
Adapters were first proposed for CNNs by \citet{Rebuffi:2017:LMV} and various formulations \cite{Pfeiffer:2021:AND,Houlsby:2019:PET,He:2022:TUV} exist for the now common ViT architecture.

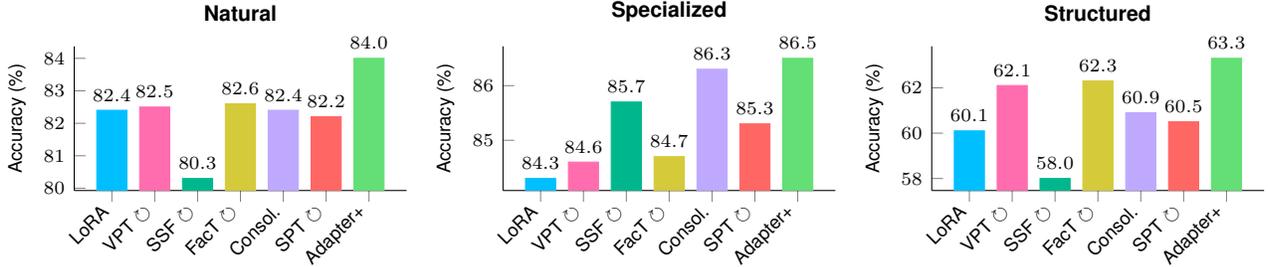
\begin{figure*}[tb]
    \centering
    \def\plotwidth{6cm}
\def\barwidth{0.4cm}

\newenvironment{barplot}[1][Title]{
    \begin{axis}[
		ybar=-\barwidth,
		axis x line*=bottom,
		axis y line*=left,
		height=3.5cm, width=\plotwidth,
		bar width=\barwidth,
		enlarge x limits={abs=0.5cm},
		ylabel={Accuracy (\%)},
		symbolic x coords={LoRA, VPT $\circlearrowright$, SSF $\circlearrowright$, FacT $\circlearrowright$, Consol., SPT $\circlearrowright$, Adapter+},
		xtick={LoRA, VPT $\circlearrowright$, SSF $\circlearrowright$, FacT $\circlearrowright$, Consol., SPT $\circlearrowright$, Adapter+},
		x tick label style={rotate=45, anchor=east, align=left},
		nodes near coords={
            \pgfmathprintnumber[zerofill,precision=1]{\pgfplotspointmeta}
        },
		nodes near coords align={vertical},
		title=\footnotesize\textbf{#1},
		font=\sffamily\scriptsize,
	]
}
{\end{axis}}

\begin{tikzpicture}
\begin{barplot}[Natural]
    \addplot[lora,fill,text=black] coordinates {(LoRA,82.4)};
	\addplot[vpt,fill,text=black] coordinates {(VPT $\circlearrowright$,82.5)};
	\addplot[ssf,fill,text=black] coordinates {(SSF $\circlearrowright$,80.3)};
	\addplot[fact,fill,text=black] coordinates {(FacT $\circlearrowright$,82.6)};
	\addplot[cons,fill,text=black] coordinates {(Consol.,82.4)};
	\addplot[spt,fill,text=black] coordinates {(SPT $\circlearrowright$,82.2)};
	\addplot[adp,fill,text=black] coordinates {(Adapter+,84.0)};
\end{barplot}

\begin{scope}[xshift=\plotwidth-0.3cm]
\begin{barplot}[Specialized]
    \addplot[lora,fill,text=black] coordinates {(LoRA,84.3)};
	\addplot[vpt,fill,text=black] coordinates {(VPT $\circlearrowright$,84.6)};
	\addplot[ssf,fill,text=black] coordinates {(SSF $\circlearrowright$,85.7)};
	\addplot[fact,fill,text=black] coordinates {(FacT $\circlearrowright$,84.7)};
	\addplot[cons,fill,text=black] coordinates {(Consol.,86.3)};
	\addplot[spt,fill,text=black] coordinates {(SPT $\circlearrowright$,85.3)};
	\addplot[adp,fill,text=black] coordinates {(Adapter+,86.5)};
\end{barplot}
\end{scope}

\begin{scope}[xshift=2*\plotwidth-0.6cm]
\begin{barplot}[Structured]
    \addplot[lora,fill,text=black] coordinates {(LoRA,60.1)};
	\addplot[vpt,fill,text=black] coordinates {(VPT $\circlearrowright$,62.1)};
	\addplot[ssf,fill,text=black] coordinates {(SSF $\circlearrowright$,58.0)};
	\addplot[fact,fill,text=black] coordinates {(FacT $\circlearrowright$,62.3)};
	\addplot[cons,fill,text=black] coordinates {(Consol.,60.9)};
	\addplot[spt,fill,text=black] coordinates {(SPT $\circlearrowright$,60.5)};
	\addplot[adp,fill,text=black] coordinates {(Adapter+,63.3)};
\end{barplot}
\end{scope}
\end{tikzpicture}   
    \vspace{-1cm}
    \caption{\textbf{Average accuracy for VTAB subgroups on the \emph{test sets}.} For methods marked with $\circlearrowright$, we report results of our re-evaluation after a complete training schedule with suitable data normalization to ensure a fair comparison. \adapter is evaluated with rank $r\!\in\![1..32]$.}
    \label{fig:vtab1k-barplots}
\end{figure*}

Recent work on parameter-efficient transfer learning \cite[\eg,][]{Jia:2022:VPT,Hao:2023:CMA,Zhang:2022:NPS,Lian:2022:SSY,He:2023:SAV,Jie:2022:FFT} presents adapters as a baseline method for the adaptation to downstream tasks in computer vision. However, we identified various common issues in their implementations, which we find to have a negative influence on the adaptation performance.
For further details, refer to the supplemental material.
Additionally, while adapters have been well studied in natural language processing (NLP), there is no study that broadly examines the different adapter configurations for vision transformers.
As a result, adapters have seemed to underperform in comparison to recent parameter-efficient adaptation methods, \eg, reported accuracies of adapters on VTAB of 73.9\% in \cite{Zhang:2022:NPS} and 60.8\% in \cite{Jia:2022:VPT}.

In this work, we therefore revisit the idea of adapters and investigate how they can perform at their best in connection with ViTs. 
Our contribution hereby is threefold:
\begin{enumerate*}[label=\emph{(\arabic*)}]
    \item We show the \emph{first in-depth and systematic study} on the effects of the adapter position in the transformer and of the adapter's inner structure with ViTs, as well as evaluate different variants of parameter initialization.
    \item We further propose a \emph{learnable, channel-wise scaling} as extension to plain adapters, which proves to be beneficial for computer vision tasks.
    \item Finally, we present \adapter, an adapter configuration with an \emph{excellent parameter-accuracy trade-off} compared to other work, as shown in \cref{fig:vtab_params_accuracy}.
    \adapter reaches a state-of-the-art average accuracy of 77.6\% on VTAB \cite{Zhai:2020:LSR} \emph{without any hyperparameter optimization per task} and 3.7 percentage points (pp) over previous adapter baselines.
    We also reach a state-of-the-art accuracy of 90.7\% on FGVC \cite{Jia:2022:VPT} with the lowest number of parameters compared to other methods.
    Finally, \adapter shows the best robustness in terms of accuracy across the VTAB subgroups, see \cref{fig:vtab1k-barplots}.
\end{enumerate*}

\section{Related work}
One possibility to adapt a pre-trained network to a novel task, apart from full fine-tuning, is to only selectively tune some of the parameters, \eg, only training the classifier \cite{Donahue:2014:DDC}. \citet{Cai:2020:TRM} proposed to tune only the biases of an otherwise frozen network to adapt it to a downstream task. BitFit \cite{Zaken:2022:BSP} then showed the efficacy of this method for NLP transformers.

\inparagraph{Modular adaptation.}
The concept of adding small, trainable modules with only a few parameters to an otherwise frozen network was first proposed for adapting CNNs by \citet{Rebuffi:2017:LMV} and called adapters.
Other approaches replaced all convolutions in the network with depth-wise separable convolutions and only tuned their spatial parts \cite{Guo:2019:DCA}, learned binary masks to prune a pre-trained network per target task \cite{Mallya:2018:PAS}, or created a student network by augmenting the original network with adapter-like modules and skip connections, which then mimicked a teacher network by disabling parts of its pre-trained and added modules \cite{Morgado:2019:NTA}.

Following the rise of transformers in NLP \cite{Vaswani:2017:AAY,Devlin:2019:BPD,Radford:2018:ILU}, \citet{Houlsby:2019:PET} proposed adapter modules in the form of bottlenecks for transformer layers.
\citet{Pfeiffer:2021:AND} conducted an architecture search on NLP tasks to find a more parameter-efficient configuration of adapter modules that only acts on the transformer's feed-forward network (FFN), thus saving roughly half of the parameters over \cite{Houlsby:2019:PET}.

\inparagraph{Prompt tuning.}
Inspired by changing the output of a network for NLP with hand-crafted textual prompts, which modifies the attention over the original input tokens, \citet{Lester:2021:PSP} proposed prompt tuning:
A set of learnable tokens is added to the input sequence and trained with backpropagation to prompt a frozen language model to perform downstream tasks.
\citet{Li:2020:PTO} extended on prompt tuning by adding learnable tokens at every transformer layer of the model, which they termed prefix tuning.
\citet{Jia:2022:VPT} applied prompt tuning to vision transformers, then called visual prompt tuning (VPT), by preprending the sequence of image patch embeddings with such trainable tokens (VPT-Shallow).
They also showed a variant resembling prefix tuning with stronger adaptation capabilities that adds tokens at every layer of the network (VPT-Deep).

\inparagraph{Low-rank approaches.}
Also focusing on the attention part of the transformer layers, \citet{Hu:2022:LLR} proposed low-rank adaptation (LoRA) where the attention weights are updated with low-rank decomposition matrices. %
The matrices can be merged with the attention weights for inference.
The structure of LoRA is very similar to an adapter, which can be seen as a superset of LoRA acting on the transformer's FFN.
\citet{He:2022:TUV} proposed a formalism to unify LoRA, adapters, and prefix tuning \cite{Li:2020:PTO}.
It allowed them to combine the beneficial aspects of all three methods into a scaled parallel adapter (Scaled PA) for NLP tasks.
AdaptFormer \cite{Chen:2022:AAV} then applied the concept of Scaled PA to vision transformers.

\inparagraph{Other related work.}
Newer approaches for vision transformers proposed different techniques to further enhance the parameter-accuracy trade-off in adaptation.
NOAH \cite{Zhang:2022:NPS} performs an architecture search for a combination of adapters, LoRA, and VPT for each task.
SSF \cite{Lian:2022:SSY} scales and shifts the features in the network after every operation, \ie, attention, FFN, layer normalization, with task-specific, trainable modules.
\citet{Jie:2022:FFT} aggregate the weights of a ViT into a single 3D tensor.
Task-specific weight updates of this tensor are learned as a matrix decomposed into parameter-efficient factors, hence they termed their method factor-tuning (FacT).
SPT \cite{He:2023:SAV} measures the importance of the weights of a pre-trained network for a downstream task. Based on a desired parameter budget, the most important parameters are chosen for tuning and adapters or LoRA are used for weight matrices that contain enough parameters of importance. 
Consolidator \cite{Hao:2023:CMA} adapts weights in multiple orderings of channel-wise groups.
The updates for all groups are merged for efficient storage and inference.

Despite these new developments, we show that the simple concept of \emph{adapters exhibits an even better parameter-accuracy trade-off} in combination with vision transformers -- if done right and with the addition of a channel-wise scaling.

\section{Adapters for vision transformers}
\label{sec:method}

\subsection{Vision transformer basics} %
In this work, we concentrate on the parameter-efficient adaptation of vision transformers (ViT) \cite{Dosovitskiy:2021:IWW}. 
The ViT is closely modeled after the transformer model for natural language processing (NLP) proposed by \citet{Vaswani:2017:AAY}. 
A learned linear projection embeds non-overlapping and flattened patches of the input image into a sequence of $n$ tokens $\bm{x} \in \mathbb{R}^{n\times d}$, where $d$ is called the hidden dimension of the transformer. 
A positional encoding is added to the embeddings and the sequence is prepended with a trainable \texttt{[CLS]} token. 
The sequence length and the dimension of the tokens stay fixed throughout the architecture.
The sequence is sent through consecutive transformer layers that each consist of a multi-head self-attention and a feed-forward network (FFN).
For the self-attention, the tokens are projected to queries, keys, and values ($\bm{Q}$, $\bm{K}$, and $\bm{V}$) and the output of each of the $M$ attention heads is calculated as
\begin{equation}
    \text{Attention}(\bm{x})=\text{Softmax}\left(\frac{\bm{Q}(\bm{x})\bm{K}(\bm{x})\tran}{\sqrt{d'}}\right)\bm{V}(\bm{x}),
\end{equation}
with $d' = \nicefrac{d}{M}$ being the inner dimension of the head.
The FFN consists of a multilayer perceptron with two linear layers (with weights $\bm{W}_i$ and biases $\bm{b}_i$) and a GELU \cite{Hendrycks:2023:GEL} non-linearity as activation in between:
\begin{equation}
    \text{FFN}(\bm{x})=\text{GELU}(\bm{x}\bm{W}_1+\bm{b}_1)\bm{W}_2+\bm{b}_2.
\end{equation}
Both attention and FFN are employed with a preceding layer normalization (LN) \cite{Ba:2016:LN} and a skip connection and, therefore, transform an input sequence $\bm{x}$ sequentially as
\begin{subequations}
\begin{align}
    \bm{x} &\mapsto \text{Attention}(\text{LN}(\bm{x})) + \bm{x} \\
    \bm{x} &\mapsto \text{FFN}(\text{LN}(\bm{x})) + \bm{x}\,.
\end{align}
\end{subequations}
To keep the notation concise, we will omit the LNs of attention and FFN in the following; each attention and FFN is assumed to be always preceded by an LN.

\subsection{Adapters and their inner structure}
\begin{figure*}[t!]
    \centering
    \begin{subfigure}{0.1546\textwidth} %
        \centering
        \includegraphics[width=\textwidth]{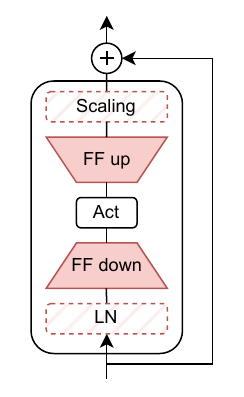}
        \caption{Inner structure}
        \label{fig:adapters-structure}
    \end{subfigure}\quad\enspace
    \begin{subfigure}{0.1371\textwidth} %
        \centering
        \includegraphics[width=\textwidth]{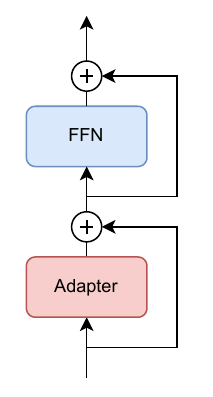}
        \caption{Pre}
        \label{fig:adapters-pre}
    \end{subfigure}\quad\enspace
    \begin{subfigure}{0.1371\textwidth} %
        \centering
        \includegraphics[width=\textwidth]{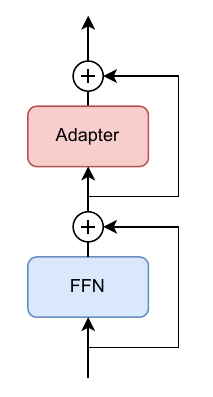}
        \caption{Post}
        \label{fig:adapters-post}
    \end{subfigure}\quad\enspace
    \begin{subfigure}{0.2138\textwidth} %
        \centering 
        \includegraphics[width=\textwidth]{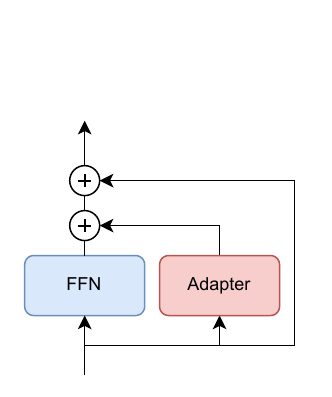}
        \caption{Parallel}
        \label{fig:adapters-parallel}
    \end{subfigure}\quad\enspace
    \begin{subfigure}{0.1546\textwidth} %
        \centering 
        \includegraphics[width=\textwidth]{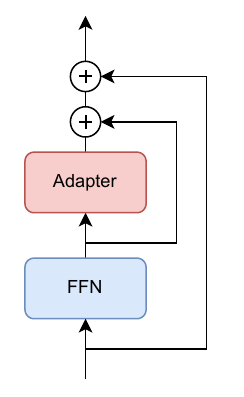}
        \caption{Intermediate}
        \label{fig:adapters-intermediate}
    \end{subfigure}
    \caption{Illustrations of (a) the \textbf{inner structure of an adapter} with feed-forward layers (FF), activation layer (Act), and optional layer normalization (LN) and scaling, (b)--(d) different possible \textbf{adapter positions} to connect the adapter to the FFN section of the transformer layer. Modules with trainable parameters are shown in \emph{red} and frozen modules in \emph{blue}.}
    \label{fig:adapters}
    \vspace{-0.5em}
\end{figure*}
Adapters \cite{Houlsby:2019:PET} are small modules that are added to the transformer layers.
They allow to tailor a network to a new task or domain, where instead of tuning the parameters of the whole network, only the adapter parameters and the classifier are trained.
Adapters take the form of bottlenecks with an inner dimension of $r \ll d$.
We call $r$ the rank of the adapter.
In detail, a down-projection to dimension $r$ with weights $\bm{W}_\text{down} \in \mathbb{R}^{d\times r}$ and biases $\bm{b}_\text{down} \in \mathbb{R}^r$ is followed by a non-linear activation function $\sigma(\cdot)$, typically a GELU \cite{Hendrycks:2023:GEL} as used throughout the ViT, and an up-projection with weights $\bm{W}_\text{up} \in \mathbb{R}^{r\times d}$ and biases ${\bm{b}_\text{up} \in \mathbb{R}^d}$ back to the hidden dimension $d$ of the transformer layer.
This yields a base adapter module
\begin{equation}
    \text{Adapter}_\text{base}(\bm{x}) = \sigma(\bm{x}\bm{W}_\text{down} + \bm{b}_\text{down}) \bm{W}_\text{up} + \bm{b}_\text{up}\,.
    \label{eq:adapter-base}
\end{equation}
The base adapter module can be further enhanced with a normalization layer, \eg, a layer normalization (LN) \cite{Ba:2016:LN}. 
Additionally, the output of the bottleneck can be scaled by $s$ as 
\begin{equation}
    \text{Adapter}(\bm{x})=s \cdot \text{Adapter}_\text{base}\big(\text{LN}(\bm{x})\big)\,.
    \label{eq:adapter}
\end{equation}
For layer-wise scaling, the factor $s$ is taken to be a scalar, \ie $s\in\mathbb{R}$, and can be either fixed as a hyperparameter or learned during training.
Layer-wise scaling was proposed by \citet{He:2022:TUV} and \citet{Hu:2022:LLR} but deemed not effective compared to a fixed scaling for tasks in NLP.
Here, we additionally propose to use a \emph{channel-wise, learned scaling} where $\bm{s} \in \mathbb{R}^d$.
We investigate its capabilities in \cref{sec:experiments-configurations}.
In most cases, the adapter is used with a skip connection, hence the complete feature transformation becomes
\begin{equation}
    \bm{x} \mapsto \text{Adapter}(\bm{x}) + \bm{x}\,.
\end{equation}
The complete inner structure of an adapter including its skip connection is visualized in \cref{fig:adapters-structure}.

\subsection{Adapter positions}
\label{sec:method-positions}
Although the architecture of bottleneck adapters for transformers is rather simple, there are various ways to plug them into the transformer layer.
Previous work has not yet investigated what the optimum position is for the use with a ViT \cite{Dosovitskiy:2021:IWW}.
Here, we evaluate four possible adapter positions, shown in \cref{fig:adapters-pre,fig:adapters-post,fig:adapters-parallel,fig:adapters-intermediate}. 
We postulate that it is easier for an adapter to learn to modify features previously transformed by a frozen module in the network rather than to anticipate what changes to the features are needed in adapting for a frozen module that follows the adapter.
Putting it differently, we argue that the adapter should follow a frozen module.

\inparagraph{Pre-Adapter.} 
The first adapter position we analyze applies the adapter to the output $\bm{x}$ of the attention section of the transformer layer before it is passed into the FFN, but with the skip connection of the attention already added (\cref{fig:adapters-pre}). 
The feature transformation of the FFN section with the adapter attached, therefore, becomes
\begin{equation}
    \label{eq:adapter-pre}
    \bm{x} \mapsto \text{FFN}\bigl(\text{Adapter}(\bm{x}) + \bm{x}\bigr) + \bigl(\text{Adapter}(\bm{x}) + \bm{x}\bigr) \,.
\end{equation}
Note that the two occurrences of $\text{Adapter}(\bm{x})$ in \cref{eq:adapter-pre} refer to the same instantiation.
In this configuration, the adapter has the full information from the feature transformation happening in the attention but needs to estimate the transformation that will be happening in the FFN that follows. 
As a result, especially the last FFN before the linear classifier will be hard to adapt. 
To the best of our knowledge, this adapter position has not been considered in the literature. %

\inparagraph{Post-Adapter.} 
In this case, the adapter is positioned at the very end of the transformer layer on the output of the FFN with its skip connection added as
\begin{equation}
    \label{eq:adapter-post}
    \bm{x} \mapsto \text{Adapter}\bigl(\text{FFN}(\bm{x}) + \bm{x}\bigr) + \bigl(\text{FFN}(\bm{x}) + \bm{x}\bigr)\,,
\end{equation}
where the FFNs refer to the same intantiation (\cref{fig:adapters-post}).
That way, the adapter has access to the feature transformation happening in the FFN and the unmodified features via the skip connection.
This position has been proposed by \citet{Pfeiffer:2021:AND} as the result of an architecture search, but only for adapting transformers for NLP tasks and not for a ViT.

\inparagraph{Parallel-Adapter.} 
Next, we consider a parallel setting as proposed by \cite{He:2022:TUV}, where the adapter is located parallel to the FFN and both share a skip connection (\cref{fig:adapters-parallel}):
\begin{equation}
    \label{eq:adapter-parallel}
    \bm{x} \mapsto \text{FFN}(\bm{x}) + \text{Adapter}(\bm{x}) + \bm{x}\,.
\end{equation}
Therefore, both adapter and FFN work on the output of the attention section of the transformer layer and the adapter needs to learn the necessary residual transformation to the one produced by the frozen FFN.

\inparagraph{Intermediate-Adapter.} 
Finally, we consider the original adapter position as proposed by \citet{Houlsby:2019:PET}. 
The adapter is plugged behind the FFN but before the skip connection of the FFN is added (\cref{fig:adapters-intermediate}). 
The adapter additionally possesses its own skip connection:
\begin{equation}
    \label{eq:adapter-intermediate}
    \bm{x} \mapsto \text{Adapter}\bigl(\text{FFN}(\bm{x})\bigr) + \text{FFN}(\bm{x}) + \bm{x}\,.
\end{equation}
Note that the two occurrences of $\text{FFN}(\bm{x})$ in \cref{eq:adapter-intermediate} refer to the same instantiation.
The adapter sees the transformed features coming from the FFN but cannot access the features added later on by the skip connection of the FFN.

\subsection{Initialization of adapter parameters}
\label{sec:method:init}

Since training a deep learning model is a non-convex optimization problem, the initialization of parameters is important. 
In this work, we evaluate three different variants of parameter initializations for adapters proposed in the literature. 
All of them have the goal to initialize the adapters in a way that minimizes the initial influence of the adapters at the start of their training.
This is a sensible goal since adapters extend an already pre-trained frozen network.

\inparagraph{Houlsby initialization.} 
\citet{Houlsby:2019:PET} propose to draw the weights of the projection matrices from a zero-centered Gaussian distribution with a standard deviation of $\sigma=0.01$, truncated at $2\sigma$, and use zero for their biases.

\inparagraph{BERT initialization.} 
For the BERT model \cite{Devlin:2019:BPD}, the initialization works similar to \citep{Houlsby:2019:PET} but the Gaussian distribution has a standard deviation of $\sigma=0.02$ and is not truncated. 
This form of initialization is used by \citet{Pfeiffer:2021:AND}.

\inparagraph{LoRA initialization.} 
LoRA \cite{Hu:2022:LLR} initializes the weights and biases of the down-projection with a uniform Kaiming He initialization \cite{He:2015:DDR}; the weights and biases of the up-projection are initialized to zero. 
Therefore, the output of the adapter at the beginning of training equals zero and the adapter initially does not contribute.

\subsection{Data normalization in pre-processing}
\label{sec:method:normalization}
Data normalization is common practice during image pre-processing. 
It is typically done by shifting and scaling of each input pixel $x_{ij}$ for each channel $c$ as
\begin{equation}
    \hat{x}_{ijc} = (x_{ijc} - \mu_c) / \sigma_c \,.
\end{equation}
Most widely used are the mean ${\bm{\mu}=(0.485, 0.456, 0.406)\tran}$ and standard deviation ${\bm{\sigma}=(0.229, 0.224, 0.225)\tran}$ of the ImageNet dataset \cite{Russakovsky:2015:ILS}, commonly referred to as ImageNet normalization.
Another option is using 0.5 for every element of $\bm{\mu}$ and $\bm{\sigma}$, which is commonly referred to as Inception normalization because it is used for the Inception family of CNN architectures, starting with Inception-v3 \cite{Szegedy:2016:RIA}. 
The ImageNet normalization aims to center the input data around $0$ with a standard deviation of $1$. %
The Inception normalization, on the other hand, transforms the input values such they are strictly in range $[-1, 1]$.

Because we try to adapt to a target domain on a very low parameter budget, it is important to use the data normalization the network saw during its pre-training. 
Otherwise, the parameter-efficient transfer method of choice needs to first compensate for the shift in input data statistics and loses parts of its capacity to adapt to the target domain.

\section{Experiments}
\label{sec:experiments}

\subsection{Datasets}
\label{sec:experiments-data}
In order to carry out a detailed study of the utility of adapters in the context of ViT models, we experiment with two standard benchmarks for task adaptation.

\inparagraph{VTAB.}
The Visual Task Adaptation Benchmark (VTAB) \cite{Zhai:2020:LSR} consists of 19 tasks, which are further grouped into three categories: 
Natural, Specialized, and Structured. 
The \emph{Natural} group contains natural images captured using standard photographic equipment. 
The \emph{Specialized} group is built from datasets of images captured with specialized equipment, from remote sensing and medical domains. 
Lastly, the \emph{Structured} group is for evaluating the understanding of the scene structure. 
Here, the majority of the datasets are compiled from synthetic images with scenes that are easy to assess for humans but have a large domain gap to natural image datasets.
Each task of VTAB consists of 800 training and 200 validation images. 
The test sets have the same number of images as the test sets in the original datasets.

\inparagraph{FGVC.}
Following \citet{Jia:2022:VPT}, we compile five datasets for fine-grained visual classification (FGVC): 
CUB-200-2011 \cite{Wah:2011:CBD}, NABirds \cite{Horn:2015:BBR}, Oxford Flowers \cite{Nilsback:2008:AFC}, Stanford Dogs \cite{Khosla:2011:NDF}, and Stanford Cars \cite{Gebru:2017:FGC}.
Because VTAB benchmarks task adaptation in a low-data regime in terms of the number of available training images, we use FGVC to evaluate adaptation methods in settings where training data is abundant.
Where validation sets are not available in FGVC, we follow \citet{Jia:2022:VPT} to create the validation splits.

\smallskip
For further details regarding the dataset properties of VTAB and FGVC, see supplemental material. %

\subsection{Experimental settings}
\label{sec:experiments-setting}
For all our experiments, we use a ViT-B/16 network \cite{Dosovitskiy:2021:IWW} that was pre-trained on ImageNet-21k \cite{Russakovsky:2015:ILS}. 
We follow its pre-training settings, in particular, regarding input data normalization. 
We train all models with an AdamW \cite{Loshchilov:2019:DWD} optimizer with a learning rate of \num{e-3}, a weight decay of \num{e-4}, and a batch size of 64, following \cite{Zhang:2022:NPS}. 
For full fine-tuning, we use a learning rate of \num{e-4}, which we found leads to better results. 
We use a cosine learning rate schedule with a linear warm-up over the first 10 epochs and train for 100 epochs in total.
We use stochastic depth with linearly increasing drop rates as a function of network depth from 0 to 0.1 for the frozen network and with a drop rate of 0.1 for the adapters during training.
Apart from data normalization (\cf \cref{sec:method:init}), we resize input images to 224$\times$224\,px for VTAB and use a randomly resize crop to 224$\times$224\,px and horizontal flipping for FGVC. 
For the ablations and to determine hyperparameters, we evaluate on the validation splits.
We include the validation sets in the training data for producing final results.

\subsection{Exploring adapter configurations}
\label{sec:experiments-configurations}

\paragraph{Adapter position.}
\begin{table}[t]
    \tabstyle{3pt}  %
    \caption{\textbf{Adapter position.} We report the average accuracy in \% ($\pm$~std.\ dev.) on the VTAB \emph{val sets} for different adapter positions. Adapter$_\text{base}$ with Houlsby initialization and rank $r\!=\!8$ is used in all experiments.}
    \label{tbl:adapter_position}
    \footnotesize
    \vspace{-0.3em}
    \begin{tabularx}{\columnwidth}{@{}Xccc@{\enspace\quad}c@{}}
\toprule
Position & Natural & Specialized & Structured & Average \\
\midrule
Pre & \underline{82.4} $\pm$ 0.4 & \textbf{86.2} $\pm$ 0.8 & 57.5 $\pm$ 0.5 & 75.3 $\pm$ 0.3 \\
Intermediate & \textbf{83.0} $\pm$ 0.4 & 85.0 $\pm$ 0.8 & 57.2 $\pm$ 0.5 & 75.1 $\pm$ 0.3 \\
Parallel & \textbf{83.0} $\pm$ 0.3 & \textbf{86.2} $\pm$ 0.6 & \underline{57.7} $\pm$ 0.6 & \underline{75.6} $\pm$ 0.3 \\
Post & \textbf{83.0} $\pm$ 0.3 & \underline{85.7} $\pm$ 0.4 & \textbf{59.1} $\pm$ 0.3 & \textbf{76.0} $\pm$ 0.2 \\
\bottomrule
\end{tabularx}

    \vspace{-0.2em}
\end{table}
We first evaluate the four possible positions to connect an adapter to the FFN section of the transformer layer, as described in \cref{sec:method-positions}. 
In our ablation, we use Adapter$_\text{base}$ (\cf \cref{eq:adapter-base}) with rank $r\!=\!8$ and use the Houlsby initialization. 
In this experiment, the adapters neither have a layer normalization nor use scaling.

The results on the VTAB validation set for all four adapter positions are presented in \cref{tbl:adapter_position}.
The \emph{Post-Adapter yields the best result} with 76.0\% average accuracy over all VTAB subgroups.
It confirms our hypothesis that the adapter should follow the frozen FFN module because it can then post-hoc modify the features flowing through the network.
The parallel configuration comes in second with 75.6\% average accuracy, receiving the same input as the FFN but having to learn a residual modification to the FFN instead of a subsequent one.
Pre-Adapter and Intermediate-Adapter are subpar compared to the other positions.
They either do not have access to the feature transformation happening afterwards in the FFN or to the features of the skip connection containing the output of the attention.

\begin{table}[t]
    \tabstyle{2pt}  %
    \caption{\textbf{Inner adapter structure.} We evaluate the different components of the adapter structure, \eg, normalization layer (\emph{Norm}), \emph{layer}-wise and \emph{channel}-wise learnable scaling on the VTAB \emph{val sets}. The difference to Adapter$_\text{base}$ (\emph{first row}) is shown in $\Delta_\text{base}$.}
    \label{tbl:adapter_structure}
    \footnotesize
    \vspace{-0.3em}
    \begin{tabularx}{\columnwidth}{@{}c@{\quad}c@{\quad}c@{\quad}cXc@{\enspace\quad}S[table-format=+1.1]@{}}
\toprule
Bias & Norm & Scaling & Initialization & & Accuracy (\%) & $\Delta_\text{base}$ \\
\midrule
\cmark &        &         & Houlsby & & 76.0 $\pm$ 0.2 & 0.0 \\
       &        &         & Houlsby & & 75.6 $\pm$ 0.4 & -0.4 \\
\cmark &        &         & LoRA    & & 75.5 $\pm$ 0.3 & -0.5 \\
\cmark &        &         & BERT    & & 75.8 $\pm$ 0.3 & -0.2 \\
\cmark & \cmark &         & Houlsby & & 75.9 $\pm$ 0.3 & -0.1 \\
\cmark & \cmark & layer   & Houlsby & & 75.9 $\pm$ 0.3 & -0.1 \\
\cmark &        & layer   & Houlsby & & \underline{76.2} $\pm$ 0.3 & \underline{+0.2}\\
\cmark & \cmark & channel & Houlsby & & 75.8 $\pm$ 0.3 & -0.2 \\
\cmark &        & channel & Houlsby & & \textbf{76.5} $\pm$ 0.2 & \textbf{+0.5} \\
\bottomrule
\end{tabularx}

    \vspace{-0.2em}
\end{table}

\inparagraph{Inner structure.}
Next, we investigate the impact of the inner structure of adapters including their initialization. 
\cref{tbl:adapter_structure} shows our findings with average accuracies calculated over the three VTAB subgroups. 
Removing the biases from the linear layers leads to a decrease in accuracy of 0.4 percentage points (pp). 
We find that the \emph{Houlsby initialization of the adapter parameters is best} while BERT and LoRA initializations reduce the accuracy by \SI{0.2}{\pp} and \SI{0.5}{\pp}.
Adding layer normalization (LN) to the adapter is slightly detrimental for all settings, both with scaling and without, while additionally adding $2d$ parameters per layer. 
We find that \emph{a learned scaling is in general beneficial} for image-classification tasks. 
Adding layer-wise scaling leads to a gain of \SI{0.2}{\pp}. 
The inclusion of a learned, channel-wise scaling, as proposed here, gives the strongest improvement of \SI{0.5}{\pp}, reaching an accuracy of 76.5\% on the VTAB validation set while only adding half of the parameters compared to LN.

\inparagraph{What makes a great adapter?}
From our systematic exploration of possible adapter configurations, we conclude that adapter modules in the \textbf{Post-Adapter} position with a learnable, \textbf{channel-wise scaling} and \textbf{Houlsby initialization} work best for computer vision tasks.
We call our proposed adapter configuration \textbf{\adapter}.
The addition of layer normalization, as suggested by \citet{Pfeiffer:2021:AND}, is not necessary and even leads to detrimental effects in our setting.

\inparagraph{Configurations from previous work.}
\begin{table}[t]
    \tabstyle{2pt}  %
    \caption{\textbf{Comparison of \adapter with adapter configurations from previous work.} We report the average accuracy in \% ($\pm$~std.\ dev.) of each subgroup and across all groups on the VTAB \emph{val sets}.} %
    \label{tbl:configurations}
    \scriptsize
    \begin{tabularx}{\columnwidth}{@{}Xc@{\enspace}ccc@{\;\quad}c@{}}
\toprule
Configuration & \#\,Param\,\tiny{(M)} & Natural & Specialized & Structured & Average \\
\midrule
Houlsby \cite{Houlsby:2019:PET}, $r\!=\!8$ & 0.39 & \underline{82.9} $\pm$ 0.2 & 85.5 $\pm$ 0.3 & \underline{58.9} $\pm$ 0.8 & \underline{75.8} $\pm$ 0.3\\
Houlsby \cite{Houlsby:2019:PET}, $r\!=\!4$ & 0.24 & \underline{82.9} $\pm$ 0.4 & 84.9 $\pm$ 0.3 & 58.3 $\pm$ 0.6 & 75.4 $\pm$ 0.3 \\
Pfeiffer \cite{Pfeiffer:2021:AND}     & 0.21 & \underline{82.9} $\pm$ 0.3 & \underline{86.1} $\pm$ 0.9 & 58.4 $\pm$ 0.7 & \underline{75.8} $\pm$ 0.4 \\
AdaptFormer \cite{Chen:2022:AAV}        & \textbf{0.19} & \textbf{83.0} $\pm$ 0.4 & 85.0 $\pm$ 0.2 & 57.4 $\pm$ 0.5 & 75.2 $\pm$ 0.2 \\
\adapter                            & \underline{0.20} & \textbf{83.0} $\pm$ 0.2 & \textbf{86.8} $\pm$ 0.6 & \textbf{59.7} $\pm$ 0.4 & \textbf{76.5} $\pm$ 0.2 \\
\bottomrule
\end{tabularx}

    \vspace{-0.5em}
\end{table}
Different configurations of adapters have been established in previous work. 
We compare their configurations to our systematic approach with rank $r\!=\!8$ on the VTAB validation sets.
Using our own implementations already leads to better results than reported in literature but enables us to compare on equal footing.
\citet{Houlsby:2019:PET} use an Intermediate-Adapter with their proposed initialization both at the FFN section as well at the attention part of the transformer layer. Additionally, they adapt the LN parameters of the backbone.
We, therefore, compare their setting additionally with $r\!=\!4$ to compare on roughly the same parameter budget.
\citet{Pfeiffer:2021:AND} suggest a Post-Adapter like us but with a BERT initialization and they employ a layer normalization inside the adapter.
AdaptFormer \cite{Chen:2022:AAV} has the same configuration as a scaled parallel adapter (Scaled PA) \cite{He:2022:TUV}, which was proposed for NLP tasks, the only difference being the layer-wise scaling~$s$.
Scaled PA uses a fixed scaling of $s=4$ for the adapters whereas AdaptFormer suggests to use $s=0.1$ for vision tasks. 
Optimizing $s$ for VTAB may lead to better results.
Our results are presented in \cref{tbl:configurations}. 
We see a clear advantage of our \adapter configuration, gaining at least \SI{0.7}{\pp} over all previous adapter realizations considered despite having the second lowest number of trainable parameters. 

\begin{table*}[t]
    \tabstyle{1pt}  %
    \caption{\textbf{Detailed results on the VTAB \emph{test sets}}. We report original results and re-evaluations ($\circlearrowright$) in \% after a complete training schedule with suitable data normalization. Grayed out numbers are not included in the ranking for \textbf{best} and \underline{second} best results. $\dagger$: Early-stopping based on the \emph{test set}, $\bullet$: unsuitable data normalization, \Lightning: per-task hyperparameter optimization. $^1$Average across the average accuracies of the VTAB groups, following previous work. $^2$No complete code release for Consolidator, hence training and evaluation details are unknown.}
    \label{tbl:vtab1k_results}
    \scriptsize
    \begingroup
\renewcommand*{\arraystretch}{0.95}
\begin{tabularx}{\linewidth}{@{}+X-c@{\enspace\quad}-c@{\enspace}-c@{\enspace}-c@{\enspace}-c@{\enspace}-c@{\enspace}-c@{\enspace}-c@{\quad}-c@{\quad\enspace}-c@{\enspace}-c@{\enspace}-c@{\enspace}-c@{\quad}-c@{\quad\enspace}-c@{\enspace}-c@{\enspace}-c@{\enspace}-c@{\enspace}-c@{\enspace}-c@{\enspace}-c@{\enspace}-c@{\quad}-c@{\quad\enspace}-c@{}}
\toprule
    & & \multicolumn{8}{c@{\enspace\quad}}{\textbf{Natural}} & \multicolumn{5}{c@{\enspace\quad}}{\textbf{Specialized}} & \multicolumn{9}{c@{\enspace\quad}}{\textbf{Structured}} \\
    \cmidrule(lr{2em}){3-10} \cmidrule(lr{2em}){11-15} \cmidrule(lr{2em}){16-24} \\ \addlinespace[-1em]
    & \rotatebox{90}{\#\,Param (M)} & \rotatebox{90}{Cifar100 \cite{Krizhevsky:2009:LML}} & \rotatebox{90}{Caltech101 \cite{Fei-Fei:2006:OSL}} & \rotatebox{90}{DTD \cite{Cimpoi:2014:DTW}} & \rotatebox{90}{Flower102 \cite{Nilsback:2008:AFC}} & \rotatebox{90}{Pets \cite{Parkhi:2012:CAD}} & \rotatebox{90}{SVHN \cite{Netzer:2011:RDN}}  & \rotatebox{90}{Sun397 \cite{Xiao:2010:SDL}} & \rotatebox{90}{Average} & \rotatebox{90}{Camelyon \cite{Veeling:2018:REC}} & \rotatebox{90}{EuroSAT \cite{Helber:2019:END}}   & \rotatebox{90}{Resisc45 \cite{Cheng:2017:RSI}}  & \rotatebox{90}{Retinopathy \cite{Dugas:2015:DRD}} & \rotatebox{90}{Average} & \rotatebox{90}{Clevr-Count \cite{Johnson:2017:CDD}} & \rotatebox{90}{Clevr-Dist. \cite{Johnson:2017:CDD}}  & \rotatebox{90}{DMLab \cite{Beattie:2016:DML}} & \rotatebox{90}{KITTI-Dist. \cite{Geiger:2013:VMR}}  & \rotatebox{90}{dSpr-Loc. \cite{Matthey:2017:DDT}} & \rotatebox{90}{dSpr-Ori \cite{Matthey:2017:DDT}}   & \rotatebox{90}{sNORB-Azi. \cite{LeCun:2004:LMG}}  & \rotatebox{90}{sNORB-Ele. \cite{LeCun:2004:LMG}} & \rotatebox{90}{Average} & \rotatebox{90}{Global Average$^1$}   \\
\midrule
    Full & \textcolor{\tblgray}{85.8} & 73.2 & 92.6 & 70.4 & 97.9 & 86.2 & 90.6 & 39.6 & 78.6 & 87.1 & 96.6 & \underline{87.5} & 74.0 & \underline{86.3} & 66.6 & 61.0 & 49.8 & 79.7 & 82.6 & 51.9 & 33.5 & 37.0 & 57.8 & 74.2 \\
    Linear & \textcolor{\tblgray}{0.04} & 78.1 & 88.1 & 69.0 & 99.1 & 90.0 & 36.0 & 56.9 & 73.9 & 79.8 & 90.7 & 73.7 & 73.7 & 79.5 & 32.4 & 30.5 & 35.9 & 61.9 & 11.2 & 26.2 & 14.3 & 24.5 & 29.6 & 61.0 \\
\midrule
    LoRA \cite{Hu:2022:LLR} & 0.29 & 83.0 & 91.7 & 71.6 & 99.2 & 90.9 & 83.8 & 56.7 & 82.4 & 86.2 & 95.7 & 83.5 & 71.9 & 84.3 & 77.7 & 62.3 & 49.0 & 80.2 & 82.2 & 51.7 & 31.0 & 47.0 & 60.1 & 75.6 \\
    \grayrow VPT-Deep \Lightning$\bullet$ \cite{Jia:2022:VPT} & 0.60 & 78.8 & 90.8 & 65.8 & 98.0 & 88.3 & 78.1 & 49.6 & 78.5 & 81.8 & 96.1 & 83.4 & 68.4 & 82.4 & 68.5 & 60.0 & 46.5 & 72.8 & 73.6 & 47.9 & 32.9 & 37.8 & 55.0 & 72.0 \\
    VPT-Deep \Lightning$\circlearrowright$ & 0.60 & 83.0 & 93.0 & 71.2 & 99.0 & 91.3 & 84.1 & 56.0 & 82.5 & 84.9 & 96.6 & 82.5 & 74.5 & 84.6 & 77.5 & 58.7 & 49.7 & 79.6 & 86.2 & \underline{56.1} & \textbf{37.9} & \textbf{50.7} & 62.1 & 76.4 \\
    NOAH \Lightning${\dagger\circbullet}$ \cite{Zhang:2022:NPS} & 0.43 & 69.6 & 92.7 & 70.2 & 99.1 & 90.4 & 86.1 & 53.7 & 80.2 & 84.4 & 95.4 & 83.9 & \underline{75.8} & 84.9 & 82.8 & \textbf{68.9} & 49.9 & \underline{81.7} & 81.8 & 48.3 & 32.8 & 44.2 & 61.3 & 75.5 \\
    \grayrow SSF \Lightning$\dagger$ \cite{Lian:2022:SSY} & 0.24 & 69.0 & 92.6 & 75.1 & 99.4 & 91.8 & 90.2 & 52.9 & 81.6 & 87.4 & 95.9 & 87.4 & 75.5 & 86.6 & 75.9 & 62.3 & 53.3 & 80.6 & 77.3 & 54.9 & 29.5 & 37.9 & 59.0 & 75.7 \\
    SSF \Lightning$\circlearrowright$ & 0.24 & 61.9 & 92.3 & 73.4 & \textbf{99.4} & \textbf{92.0} & \underline{90.8} & 52.0 & 80.3 & 86.5 & 95.8 & \underline{87.5} & 72.8 & 85.7 & 77.4 & 57.6 & 53.4 & 77.0 & 78.2 & 54.3 & 30.3 & 36.1 & 58.0 & 74.6 \\
    \grayrow FacT-TK$_{8}$ \Lightning$\dagger\bullet$ \cite{Jie:2022:FFT} & 0.05 & 70.3 & 88.7 & 69.8 & 99.0 & 90.4 & 84.2 & 53.5 & 79.4 & 82.8 & 95.6 & 82.8 & 75.7 & 84.2 & 81.1 & 68.0 & 48.0 & 80.5 & 74.6 & 44.0 & 29.2 & 41.1 & 58.3 & 74.0 \\
    $\text{FacT-TK}_{8}$ \Lightning$\circlearrowright$ & \textbf{0.05} & 74.9 & 92.7 & 73.7 & 99.1 & 91.3 & 85.5 & 57.7 & 82.1 & 86.8 & 94.9 & 84.1 & 70.9 & 84.2 & 81.9 & 64.1 & 49.2 & 77.2 & 83.8 & 53.1 & 28.2 & 44.7 & 60.3 & 75.5 \\
    \grayrow FacT-TK$_{\leq 32}$ \Lightning$\dagger\bullet$ \cite{Jie:2022:FFT} & 0.10 & 70.6 & 90.6 & 70.8 & 99.1 & 90.7 & 88.6 & 54.1 & 80.6 & 84.8 & 96.2 & 84.5 & 75.7 & 85.3 & 82.6 & 68.2 & 49.8 & 80.7 & 80.8 & 47.4 & 33.2 & 43.0 & 60.7 & 75.6 \\
    FacT-TK$_{\leq 32}$ \Lightning$\circlearrowright$ & 0.10 & 74.6 & 93.7 & \underline{73.6} & \underline{99.3} & 90.6 & 88.7 & 57.5 & 82.6 & 87.6 & 95.4 & 85.5 & 70.4 & 84.7 & \textbf{84.3} & 62.6 & 51.9 & 79.2 & 85.5 & 52.0 & 36.4 & 46.6 & 62.3 & 76.5 \\
    Consolidator$\,^2$ \cite{Hao:2023:CMA} & 0.30 & 74.2 & 90.9 & \textbf{73.9} & \textbf{99.4} & \underline{91.6} & \textbf{91.5} & 55.5 & 82.4 & 86.9 & 95.7 & 86.6 & \textbf{75.9} & \underline{86.3} & 81.2 & \underline{68.2} & 51.6 & \textbf{83.5} & 79.8 & 52.3 & 31.9 & 38.5 & 60.9 & 76.5 \\
    \grayrow SPT-Adapter ${\dagger\bullet}$ \cite{He:2023:SAV} & 0.23 & 72.9 & 93.2 & 72.5 & 99.3 & 91.4 & 84.6 & 55.2 & 81.3 & 85.3 & 96.0 & 84.3 & 75.5 & 85.3 & 82.2 & 68.0 & 49.3 & 80.0 & 82.4 & 51.9 & 31.7 & 41.2 & 60.8 & 75.8 \\
    SPT-Adapter $\circlearrowright$ & 0.22 & 74.7 & \underline{94.1} & 73.0 & 99.1 & 91.2 & 84.5 & 57.5 & 82.0 & 85.7 & 94.9 & 85.7 & 70.2 & 84.1 & 81.3 & 63.2 & 49.1 & 80.7 & 83.5 & 52.0 & 26.4 & 41.5 & 59.7 & 75.3 \\
    \grayrow SPT-Adapter ${\dagger\bullet}$ \cite{He:2023:SAV} & 0.43 & 72.9 & 93.2 & 72.5 & 99.3 & 91.4 & 88.8 & 55.8 & 82.0 & 86.2 & 96.1 & 85.5 & 75.5 & 85.8 & 83.0 & 68.0 & 51.9 & 81.2 & 82.4 & 51.9 & 31.7 & 41.2 & 61.4 & 76.4 \\
    SPT-Adapter $\circlearrowright$ & 0.43 & 74.9 & 93.2 & 71.6 & 99.2 & 91.1 & 87.9 & 57.2 & 82.2 & 87.0 & 95.4 & 86.5 & 72.4 & 85.3 & 81.1 & 63.2 & 50.3 & 80.2 & 84.4 & 51.4 & 31.5 & 42.2 & 60.5 & 76.0 \\
    
\midrule
    \adapter, $r\!=\!1$ & \underline{0.07} & \textbf{85.4} & 92.4 & 73.1 & 99.1 & 91.3 & 83.1 & \textbf{58.1} & 83.2 & 87.2 & 96.6 & 85.3 & 72.6 & 85.5 & 80.7 & 60.6 & 50.9 & 79.9 & 83.3 & 55.6 & 27.1 & 43.0 & 60.1 & 76.3 \\
    \adapter, $r\!=\!2$ & 0.09 & \textbf{85.4} & 93.0 & 72.7 & 99.2 & 90.6 & 85.3 & \underline{58.0} & 83.5 & \textbf{87.9} & 96.8 & 85.5 & 71.4 & 85.4 & 83.2 & 61.0 & 51.6 & 80.1 & 86.1 & \textbf{56.3} & 30.7 & 46.5 & 61.9 & 76.9 \\
    \adapter, $r\!=\!4$ & 0.13 & \underline{84.8} & 93.8 & 72.7 & 99.2 & 90.6 & 86.5 & 57.4 & 83.6 & 87.5 & \underline{96.9} & 85.9 & 71.5 & 85.4 & \underline{83.4} & 61.6 & 53.6 & 81.4 & 87.3 & 55.3 & 34.4 & \underline{48.1} & \underline{63.1} & 77.4 \\
    \adapter, $r\!=\!8$ & 0.20 & 84.6 & \textbf{94.2} & 72.3 & \underline{99.3} & 90.7 & 87.6 & 56.7 & 83.6 & \underline{87.7} & \textbf{97.0} & 86.7 & 72.3 & 85.9 & 83.2 & 60.9 & \textbf{53.8} & 80.3 & \underline{88.1} & 55.6 & 35.7 & 47.7 & \underline{63.1} & 77.6 \\
    \adapter, $r\!=\!16$ & 0.35 & 83.7 & \textbf{94.2} & 71.5 & \underline{99.3} & 90.6 & 88.2 & 55.8 & 83.3 & 87.5 & \textbf{97.0} & 87.4 & 72.9 & 86.2 & 82.9 & 60.9 & \underline{53.7} & 80.8 & \textbf{88.4} & 55.2 & \underline{37.3} & 46.9 & \textbf{63.3} & 77.6 \\
\midrule
    \adapter, $r\!\in\![1..4]$ \Lightning & 0.11 & \textbf{85.4} & 93.8 & 72.7 & 99.1 & 90.6 & 86.5 & \textbf{58.1} & 83.7 & 87.5 & 96.8 & 85.9 & 71.4 & 85.4 & \underline{83.4} & 61.0 & 53.6 & 81.4 & 87.3 & 55.3 & 34.4 & \underline{48.1} & \underline{63.1} & 77.4 \\
    \adapter, $r\!\in\![1..8]$ \Lightning & 0.16 & \textbf{85.4} & 93.8 & 72.7 & 99.1 & 90.7 & 87.6 & \textbf{58.1} & \underline{83.9} & \underline{87.7} & 96.8 & 86.7 & 72.3 & 85.9 & \underline{83.4} & 60.9 & \textbf{53.8} & 80.3 & \underline{88.1} & 55.3 & 35.7 & 47.7 & \underline{63.1} & \underline{77.7} \\
    \adapter, $r\!\in\![1..32]$ \Lightning & 0.27 & \textbf{85.4} & 93.8 & 72.7 & 99.1 & 90.7 & 88.2 & \textbf{58.1} & \textbf{84.0} & 87.5 & 96.8 & \textbf{87.8} & 73.9 & \textbf{86.5} & \underline{83.4} & 60.9 & \textbf{53.8} & 80.3 & 87.2 & 55.3 & \textbf{37.9} & 47.7 & \textbf{63.3} & \textbf{77.9} \\
\bottomrule
\end{tabularx}
\endgroup

    \vspace{-0.9em}
\end{table*}

\subsection{Main results}
\label{sec:experiments-main}

\paragraph{VTAB.}
We evaluate \adapter on the VTAB test sets and compare to other methods in \cref{tbl:vtab1k_results}.
We provide results for \emph{full} fine-tuning and tuning only the \emph{linear} classifier while freezing the rest of the backbone \cite{Donahue:2014:DDC} as a baseline of classical fine-tuning methods.
As competing parameter-efficient tuning methods, we include LoRA \cite{Hu:2022:LLR}, VPT \cite{Jia:2022:VPT}, NOAH \cite{Zhang:2022:NPS}, SSF \cite{Lian:2022:SSY}, FacT \cite{Jie:2022:FFT}, Consolidator \cite{Hao:2023:CMA}, and SPT \cite{He:2023:SAV}.

Wherever possible, we re-evaluate the other methods with a suitable data normalization for the pre-trained backbone and after the full training schedule to enable a fair comparison.
For LoRA, we use our own implementation because the original work does not cover VTAB.
For VPT, we adopt the number of tokens per task from their hyperparameter optimization but find that we do not need to tune learning rate and weight decay per task.
Additionally, deviating from the original implementation, we optimize with AdamW \cite{Loshchilov:2019:DWD} instead of SGD~\cite{Rosenblatt:1958:PPM} and change to an appropriate data normalization.
We present the original results from \cite{Jia:2022:VPT} on VTAB together with our re-evaluation.
Our improved implementation of VPT increases the average accuracy by \SI{4.4}{\pp} from 72.0\% to 76.4\%.
SSF, FacT, and SPT released code to evaluate on VTAB.
For FacT and SPT, we change the data normalization to match the backbone; SSF already uses the correct one.
We re-run the provided code and present the results after a full training schedule.
For completeness, we also report the results from the original publications.
However, we found that the code releases of \cite{Lian:2022:SSY,Jie:2022:FFT,He:2023:SAV} use early stopping based on the best result on the \emph{test set}.
We argue that tuning hyperparameters such as the number of training epochs on the test set goes against established practices in machine learning; rather the validation set should be used for early stopping.
Yet, due to the limited size of the training and validation sets in VTAB, it is not feasible to report test results without also training on the validation data.
Hence, we chose to complete a full training schedule of 100 epochs instead of using early stopping.
Training SSF for the full schedule leads to a decrease in average accuracy of \SI{1.1}{\pp} over the original publication and re-evaluating SPT leads to a decrease of up to \SI{0.5}{\pp}, even with a corrected data normalization.
FacT on the other hand benefits from our re-revaluation, since the accuracy decrease from training a complete schedule is offset by improvements from applying the appropriate data normalization.
There was no complete code release with configurations to train Consolidator on VTAB at the time of writing, hence we report results as-is.

\adapter shows the best parameter-accuracy trade-off among all methods evaluated.
This can also be clearly seen in \cref{fig:vtab_params_accuracy}.
Additionally, \adapter sets a new state of the art with an average accuracy of up to 77.6\% over all VTAB subgroups \emph{even without any per-task hyperparameter optimization}.
If we determine the optimal rank $r$ per task on the validation set, we can further improve the accuracy to 77.9\%.
Optimizing the rank leads to a better parameter-accuracy trade-off than using a fixed rank across all tasks.

In \cref{fig:vtab1k-barplots}, we compare the average accuracy on the subgroups of VTAB. Wherever possible, we present the results of re-evaluating methods after the last training epoch and matching the data normalization to the backbone. The average accuracies of \adapter with $r\in [1..32]$ are consistently higher than those of the competing methods. Note that the accuracies of other methods except SPT differ drastically across the different VTAB subgroups. \adapter, on the other hand, shows a high degree of robustness to the domain shifts between groups.

\inparagraph{FGVC.}
\begin{table}[tb]
    \tabstyle{1pt}  %
    \caption{\textbf{Detailed results on the FGVC \emph{test sets}.} We report original results and re-evaluations ($\circlearrowright$) in \% after a complete training schedule with suitable data normalization. Grayed out numbers are not included in the ranking for \textbf{best} and \underline{second} best results.}
    \label{tbl:fgvc_results}
    \footnotesize
    \begin{tabularx}{\columnwidth}{@{}+X-c@{\enspace\quad}-c@{\enspace}-c@{\enspace}-c@{\enspace}-c@{\enspace}-c@{\enspace\quad}-c@{}}
\toprule
    & \rotatebox{90}{\#\,Param (M)} & \rotatebox{90}{CUB200 \cite{Wah:2011:CBD}} & \rotatebox{90}{NABirds \cite{Horn:2015:BBR}} & \rotatebox{90}{Oxford Flowers \cite{Nilsback:2008:AFC}} & \rotatebox{90}{Stanford Dogs \cite{Khosla:2011:NDF} }& \rotatebox{90}{Stanford Cars \cite{Gebru:2017:FGC}} & \rotatebox{90}{Average} \\
\midrule
Full & \textcolor{\tblgray}{86.0} & 88.0 & 81.5 & 99.2 & 85.6 & 90.6 & 89.0 \\
Linear & \textcolor{\tblgray}{0.18}  & 88.9 & 81.8 & \underline{99.5} & 92.6 & 52.8 & 83.1 \\
\midrule
\grayrow VPT-Deep \cite{Jia:2022:VPT} & 0.85 & 88.5 & 84.2 & 99.0 & 90.2 & 83.6 & 89.1 \\
VPT-Deep $\circlearrowright$ & 0.85 & \textbf{90.1} & 83.3 & \textbf{99.6} & 90.3 & 85.0 & 89.7 \\
\grayrow SSF \cite{Lian:2022:SSY} & 0.39 & 89.5 & 85.7 & 99.6 & 89.6 & 89.2 & 90.7 \\
SSF $\circlearrowright$ & \underline{0.39} & 88.9 & \textbf{85.0} & \textbf{99.6} & 88.9 & 88.9 & 90.3 \\
SPT-Adapter \cite{He:2023:SAV} & 0.40 & 89.1 & 83.3 & 99.2 & 91.1 & 86.2 & 89.8 \\
SPT-LoRA \cite{He:2023:SAV} & 0.52 & 88.6 & \underline{83.4} & \underline{99.5} & 91.4 & 87.3 & 90.1 \\
\midrule
\adapter, $r\!\in\![1..32]$ & \textbf{0.34} & \underline{90.0} & 83.2 & \textbf{99.6} & \underline{91.6} & \underline{89.1} & \textbf{90.7} \\
\grayrow \adapter (best epoch) & 0.34 & 90.4 & 85.0 & 99.7 & 92.6 & 89.1 & 91.4 \\
\bottomrule
\end{tabularx}

    \vspace{-0.8em}
\end{table}
Next, we present our results on the FGVC benchmark in \cref{tbl:fgvc_results}.
From the contenders, only SSF \cite{Lian:2022:SSY} has released code and hyperparameter configurations for training on FGVC at the time of writing.
As we know from the code releases for VTAB, the reported numbers show the accuracy for early stopping based on the \emph{test set}.
Therefore, we expect a similar evaluation for FGVC.
While we do not endorse early stopping based on the test set, we additionally provide numbers for that setting in \cref{tbl:fgvc_results} for the sake of comparability. 
Even when training for a complete schedule, \adapter shows the best average accuracy with 90.7\% over all five datasets in FGVC, \SI{0.4}{\pp} over the second best method under similar evaluation.
When early stopping with the test set, \adapter reaches 91.4\% average accuracy, \SI{0.7}{\pp} over the second best method and \SI{2.4}{\pp} better than full fine-tuning.
This demonstrates that \adapter also yields state-of-the-art results for task adaptation when training data is abundant while having the best parameter efficiency.

\subsection{Ablations}
\label{sec:experiments-ablations}

\paragraph{Data normalization.}
We showcase the effect of using an unsuitable data normalization for the chosen ViT in \cref{tbl:vtab1k_in_norm}. 
The gap between ImageNet and Inception normalization (see \cref{sec:method:normalization}) is largest for VPT \cite{Jia:2022:VPT}, with a \SI{3.4}{\pp} difference in average accuracy, which explains around two-thirds of the gain for our re-evaluation as shown in \cref{fig:vtab_params_accuracy}. 
We suspect that VPT has less of an ability to scale and shift the data because the learnable tokens only act on the attention mechanism. 
LoRA \cite{Hu:2022:LLR}, FacT \cite{Jie:2022:FFT}, and adapters all employ linear layers that can directly scale and shift the features of the frozen backbone and thus compensate better for improper data normalization. 
It is worth mentioning that our \adapter is the most robust to improper normalization out of the methods evaluated, with a gap of only \SI{2.6}{\pp} average accuracy.

\inparagraph{Training regularization.}
We investigate the importance of training regularization methods like stochastic depth \cite{Huang:2016:DNS} and dropout \cite{Gal:2016:DBA} for training adapters on a frozen ViT backbone and evaluate on the VTAB validation sets.
We use linearly increasing drop rates as a function of network depth from 0~to~0.1 for the frozen layers of the ViT model, and a drop rate of 0.1 when using dropout or stochastic depth for the adapter modules.
The results in \cref{tbl:adapter_drop} show a clear benefit for using stochastic regularization for the frozen layers as well as the adapters during training.
Using dropout in the adapters is only slightly better than no regularization for adapters, with a gain of only \SI{0.1}{\pp}. 
With an increase in accuracy of \SI{0.7}{\pp}, \emph{stochastic depth is the preferred regularization method for adapters}. 
However, our results show that the more important part is the \emph{stochastic depth regularization for the frozen modules of the ViT backbone}.
Disabling it in training leads to a loss of \SI{1.5}{\pp} accuracy compared to a training where stochastic depth is used throughout the model.
\begin{table}[t]
    \tabstyle{1pt}  %
    \caption{\textbf{Effects of ImageNet \vs Inception data normalization.} All methods are evaluated on the VTAB \emph{val sets}. In column $\Delta_\text{Average}$ we report the increase in accuracy in \textup{pp} across all VTAB subgroups.}
    \label{tbl:vtab1k_in_norm}
    \vspace{-1.2mm} %
    \footnotesize
    \begin{tabularx}{\linewidth}{@{}X@{\enspace}c@{\enspace}c@{\enspace}c@{\quad}c@{\enspace\quad}c@{\enspace}c@{\enspace}c@{\quad}c@{\enspace\quad}c@{}}
\toprule
 & \multicolumn{4}{c@{\enspace\quad}}{\textbf{ImageNet norm}} & \multicolumn{4}{c@{\enspace\quad}}{\textbf{Inception norm}} &  \\ 
 \cmidrule(lr{2em}){2-5} \cmidrule(lr{2em}){6-9} \\ \addlinespace[-1.8em]
 & \rotatebox{90}{Natural} & \rotatebox{90}{Specialized\ \ \ } & \rotatebox{90}{Structured} & \rotatebox{90}{Average}  & \rotatebox{90}{Natural} & \rotatebox{90}{Specialized} & \rotatebox{90}{Structured} & \rotatebox{90}{Average} & \rotatebox{90}{$\Delta_\text{Average}$} \\
\midrule
VPT & \underline{79.2} & 83.0 & 53.8 & 72.0 & \underline{82.2} & \underline{86.2} & 57.9 & \underline{75.4} & 3.4 \\
LoRA & 78.4 & \underline{84.1} & 53.2 & 71.9 & 82.0 & 85.8 & 56.4 & 74.7 & 2.8 \\
FacT-TK & 78.0 & 83.3 & \underline{56.1} & \underline{72.4} & 81.6 & 85.6 & \underline{58.1} & 75.1 & \underline{2.7} \\
\adapter & \textbf{80.5} & \textbf{85.0} & \textbf{56.0} & \textbf{73.9} & \textbf{83.0} & \textbf{86.8} & \textbf{59.7} & \textbf{76.5} & \textbf{2.6} \\
\bottomrule
\end{tabularx}

    \vspace{-1.1em}
\end{table}
\begin{table}[tb]
    \tabstyle{4pt}  %
    \caption{\textbf{Influence of training regularization.} We evaluate accuracy in \% with Adapter$_\text{base}$ with rank $r\!=\!8$ on the VTAB \emph{val sets}.}
    \label{tbl:adapter_drop}
    \vspace{-1.5mm} %
    \footnotesize
    \begin{tabularx}{\columnwidth}{@{}cXccc@{}}
\toprule
 & & \multicolumn{3}{c}{\textbf{Adapter}} \\
 \cmidrule(lr{.3em}){3-5} \\ \addlinespace[-1.2em]
 & & Stochastic Depth & Dropout & None \\
\midrule
\multirow{2}{*}{\rotatebox{90}{\textbf{ViT}}}
& Stochastic Depth & \textbf{76.0} & \underline{75.4} & 75.3 \\
& None             & 74.5 & 74.3 & 73.7 \\
\bottomrule
\end{tabularx}

    \vspace{-4.9mm}
\end{table}

\section{Conclusion}
\label{sec:conclusion}
Applied at the right position and with an optimal inner structure, the simple concept of adapters produces state-of-the-art results for task adaptation. %
To understand how adapters can ``strike back'', we conducted the first systematic and in-depth study on how to best construct adapters and integrate them with vision transformers.
This allowed us to determine the optimal connection point for the adapter in the transformer layer.
Further, we proposed to use a learnable, channel-wise scaling and showed its benefit for computer vision tasks.
Our insights led us to the creation of \adapter that yields the highest accuracy and the best parameter-accuracy trade-off on VTAB (77.6\%, 0.2M) without any per-task hyperparameter optimization and on FGVC (90.7\%,~0.34M), showing its superiority over more complicated methods.

{\small
\inparagraph{Acknowledgements.} This work has been funded by the LOEWE initiative (Hesse, Germany) within the emergenCITY center.
}

{
    \small
    \bibliographystyle{ieeenat_fullname}
    \bibliography{short,local,supp,papers,external}
}

\appendix
\maketitlesupplementary
\pagenumbering{roman}

In this appendix, we provide further details and results, which could not be included in the main paper due to space limitations.

\section{Why did adapters underperform for ViTs?}
First, we want to shed more light on why adapters do not rank well in the literature for parameter-efficient transfer learning for vision tasks.
By comparison of numbers reported for adapters on VTAB in the publications referenced in \cref*{tbl:vtab1k_results} of the main paper, we found that they essentially stem from only two sources.

The first source is VPT \cite{Jia:2022:VPT}, where results for an adapter with a reduction factor of \num{256}, amongst other configurations, are reported.
For a ViT-B/16 with a hidden dimension of $d\!=\!768$, this is equal to an adapter with rank $r\!=\!3$.
Despite citing \citet{Pfeiffer:2021:AND}, who suggest a Post-Adapter position, the actual implementation in the code base\footnote{\url{https://github.com/KMnP/vpt}} equals an Intermediate-Adapter that performs worse on VTAB (see \cref*{sec:method-positions} of the main paper).
The initialization used for the adapter parameters most resembles a LoRA initialization but sets the adapter parameters to zero everywhere.
Therefore, there is no randomization in the initialization of the adapter parameters, and different seeds only affect the initialization of the classifier.
Additionally, the intermediate features in the adapter bottlenecks then become all zero, leading to identical gradients in the up-projections at the start of training, which hinders optimization.
As a result, the adapter baseline used by VPT only reaches 60.0\% average accuracy on the VTAB test sets.
This is a gap of 17.6 percentage points (pp) compared to our \adapter with rank $r\!=\!8$ (77.6\% average accuracy).
Even when considering the loss of around \qtyrange{2}{3}{\pp} caused by an unsuitable data normalization in the VPT implementation, this is still a very significant gap.
The numbers for an adapter with rank $r\!=\!3$ from VPT are also reported in \cite{Lian:2022:SSY} as a baseline. %

The second source for adapter baseline results is the NOAH pre-print \cite{Zhang:2022:NPS}. There, an adapter with rank $r\!=\!8$ is used. Its implementation\footnote{\url{https://github.com/ZhangYuanhan-AI/NOAH}} performs the following feature transformation:
\begin{equation}
    \bm{x} \mapsto \text{Adapter}\bigl(\text{FFN}(\bm{x})\bigr) + \bm{x} \,.
\end{equation}
This is closest to the Intermediate-Adapter (\cf \cref*{eq:adapter-intermediate} of the main paper) but misses the skip connection bypassing the adapter and containing $\text{FFN}(\bm{x})$.
Thus, the adapter does not learn a residual function to an identity mapping but instead must learn a more complex mapping to transform its input.
Therefore, the adapter becomes harder to train \cite{He:2016:DRL}, leading to an average accuracy of 73.9\% on the VTAB test sets or \SI{3.7}{\pp} behind our \adapter.
For the NOAH adapter results, we see a proliferation to the publications of FacT \cite{Jie:2022:FFT} and SPT \cite{He:2023:SAV}.
The adapter implementation from NOAH is also used in the code released for Consolidator\footnote{\url{https://github.com/THU-MIG/Consolidator}} \cite{Hao:2023:CMA} but their results are produced with rank $r\!=\!16$, giving a slightly better average accuracy of 74.3\%, or \SI{3.3}{\pp} less than \adapter.

In summary, the examined baseline implementations differ from the configurations proposed by \citet{Houlsby:2019:PET} and \citet{Pfeiffer:2021:AND} and introduce issues that lead to their underperformance.
In our paper, we show that adapters are capable of reaching 77.6\% average accuracy for rank $r\!=\!8$ and 77.9\% for our optimized version of \adapter, uplifting adapters from an easy-to-beat baseline to a state-of-the-art transfer method.

\section{Dataset properties}
\label{sec:dataset_details}
In \cref{tbl:dataset_details_vtab,tbl:dataset_details_fgvc}, we show the statistics of each task in VTAB \cite{Zhai:2020:LSR} and FGVC \cite{Jia:2022:VPT} with regard to the number of classes and the number of images in the train, validation, and test splits. The tables are largely ``borrowed'' from \cite{Jia:2022:VPT}.

\begin{table}[h]
    \tabstyle{3pt}  %
    \caption{\textbf{Dataset details for VTAB.}}
    \label{tbl:dataset_details_vtab}
    \vspace{-0.3em}
    \footnotesize
    \sisetup{group-minimum-digits=4}
\begin{tabularx}{\linewidth}{@{}l@{\enspace}XS[table-format=3.0]@{\quad}c@{\enspace\ }c@{\enspace\;}S[table-format=5.0]@{}}
\toprule
\multirow{2.4}*{\textbf{Group}} & \multirow{2.4}*{\textbf{Task}} & {\multirow{2.4}*{\textbf{\#\,Classes}}} & \multicolumn{3}{c}{\textbf{Splits}} \\\cmidrule{4-6}
 & & & {Train} & {Val} & {Test} \\
\midrule
\multirow{7}{*}{Natural} & CIFAR-100 \cite{Krizhevsky:2009:LML} & 100 & \multirow{7}*{\tablenum[table-format=3.0]{800}} & \multirow{7}*{\tablenum[table-format=3.0]{200}} & 10000 \\
& Caltech-101 \cite{Fei-Fei:2006:OSL} & 102 & & & 6084 \\
& DTD \cite{Cimpoi:2014:DTW} & 47 & & & 1880 \\
& Oxford Flowers \cite{Nilsback:2008:AFC} & 102 & & & 6149 \\
& Pets \cite{Parkhi:2012:CAD} & 37 & & & 3669 \\
& SVHN \cite{Netzer:2011:RDN} & 10 & & & 26032 \\
& Sun397 \cite{Xiao:2010:SDL} & 397 & & & 21750 \\
\midrule
\multirow{4}{*}{Specialized} & Patch Camelyon \cite{Veeling:2018:REC} & 2 & \multirow{4}*{\tablenum[table-format=3.0]{800}} & \multirow{4}*{\tablenum[table-format=3.0]{200}} & 32768 \\
& EuroSAT \cite{Helber:2019:END} & 10 & & & 5400 \\
& RESISC45 \cite{Cheng:2017:RSI} & 45 & & & 6300 \\
& Diabetic Retinopathy \cite{Dugas:2015:DRD} & 5 & & & 42670 \\
\midrule
\multirow{8}{*}{Structured} & CLEVR-Count \cite{Johnson:2017:CDD} & 8 & \multirow{8}*{\tablenum[table-format=3.0]{800}} & \multirow{8}*{\tablenum[table-format=3.0]{200}} & 15000 \\
& CLEVR-Distance \cite{Johnson:2017:CDD} & 6 & & & 15000 \\
& DMLab \cite{Beattie:2016:DML} & 6 & & & 22735 \\
& KITTI-Distance \cite{Geiger:2013:VMR} & 4 & & & 711 \\
& dSprites-Location \cite{Matthey:2017:DDT} & 16 & & & 73728 \\
& dSprites-Orientation \cite{Matthey:2017:DDT} & 16 & & & 73728 \\
& smallNORB-Azimuth \cite{LeCun:2004:LMG} & 18 & & & 12150 \\
& smallNORB-Elevation \cite{LeCun:2004:LMG} & 9 & & & 12150 \\
\bottomrule
\end{tabularx}

\end{table}

\begin{table}[t]
    \tabstyle{3pt}  %
    \caption{\textbf{Dataset details for FGVC.} For datasets marked \mbox{with *}, we follow \cite{Jia:2022:VPT} to randomly sample train and validation splits because validation sets are not available from the original datasets.}
    \label{tbl:dataset_details_fgvc}
    \vspace{-0.3em}
    \footnotesize
    \sisetup{group-minimum-digits=4}
\begin{tabularx}{\linewidth}{@{}lXS[table-format=3.0]@{\qquad}S[table-format=5.0]@{\quad\ }S[table-format=4.0]@{\quad\ }S[table-format=4.0]@{}}
\toprule
\multirow{2.4}*{\textbf{Dataset}} & & {\multirow{2.4}*{\textbf{\#\,Classes}}} & \multicolumn{3}{c}{\textbf{Splits}} \\\cmidrule{4-6}
& & & {Train} & {Val} & {Test} \\
\midrule
CUB-200-2011* \cite{Wah:2011:CBD} & & 200 & 5394 & 600 & 5794 \\
NABirds* \cite{Horn:2015:BBR} & & 555 & 21536 & 2393 & 6084 \\
Oxford Flowers \cite{Nilsback:2008:AFC} & & 102 & 1020 & 1020 & 6149 \\
Stanford Dogs* \cite{Khosla:2011:NDF} & & 120 & 10800 & 1200 & 8580 \\
Stanford Cars* \cite{Gebru:2017:FGC} & & 196 & 7329 & 815 & 8041 \\
\bottomrule
\end{tabularx}

    \vspace{-1.2em}
\end{table}

\section{More experimental settings}
For all experiments conducted with our implementation, we average the results over three seeds.
This includes the (re\nobreakdash-)evaluations of LoRA and VPT.
We built our implementation on PyTorch \cite{Paszke:2019:PIS}, PyTorch Lightning,\footnote{\url{https://lightning.ai/pytorch-lightning}} and timm.\footnote{\url{https://github.com/huggingface/pytorch-image-models}}
We run experiments with bfloat16 mixed precision on a NVIDIA RTX A6000 GPU.

For our experiments in the main paper, we report results for a fixed adapter rank $r$ as well as ranks optimized per task.
For the per-task optimization of \adapter, we use a hyperparameter sweep over the set of ranks $r\!\in\!\{1,2,4,8,16,32\}$.
We evaluate on the validation sets of VTAB and FGVC and choose the per-task ranks from the specified range(s) to steer the number of average parameters.
The ranks we used to produce the results on the VTAB and FGVC test sets (see \cref*{tbl:vtab1k_results,tbl:fgvc_results} in the main paper) are shown in detail in \cref{tbl:adapter_ranks_vtab} and \cref{tbl:adapter_ranks_fgvc}, respectively.

\section{Calculation of no.~of trainable parameters}
Suppose we have a ViT with a hidden dimension $d$, $N$ transformer layers, and adapters with rank $r$.
The total number of learnable parameters for Adapter$_\text{base}$ modules (\cf \cref*{eq:adapter-base} of the main paper) attached to the FFN of every transformer layer then amounts to $N (2dr + r + d)$.
Including layer normalization in the adapter modules amounts to $N 2d$ additional parameters.
The addition of learned, layer-wise scaling amounts to $N$ extra parameters and choosing learned, channel-wise scaling instead adds $Nd$ extra parameters.
\adapter (see \cref*{sec:experiments-configurations} of the main paper) thus amounts to $N(2dr+2d+r)$ total parameters.
Additionally, for a task with $c$ classes, we add a classifier with $dc+c$ learnable parameters.

\begin{table}[t]
    \tabstyle{1pt}  %
    \caption{\textbf{Adapter rank $r$ for each VTAB task} for optimized versions of \adapter with different ranges of permitted ranks.}
    \label{tbl:adapter_ranks_vtab}
    \vspace{-0.3em}
    \scriptsize
    \begingroup
\begin{tabularx}{\linewidth}{@{}Xc@{\quad}c@{\:\,}c@{\:\,}c@{\:\,}c@{\:\,}c@{\:\,}c@{\:\,}c@{\quad}c@{\:\,}c@{\:\,}c@{\:\,}c@{\quad}c@{\:\,}c@{\:\,}c@{\:\,}c@{\:\,}c@{\:\,}c@{\:\,}c@{\:\,}c@{}}
\toprule
    & & \multicolumn{7}{c@{\quad}}{\textbf{Natural}} & \multicolumn{4}{c@{\quad}}{\textbf{Specialized}} & \multicolumn{8}{c}{\textbf{Structured}} \\
    \cmidrule(l{.2em}r{1em}){3-9} \cmidrule(l{.2em}r{1em}){10-13} \cmidrule(l{.2em}r{.2em}){14-21} \\ \addlinespace[-.9em]
    & \rotatebox{90}{\#\,Param (M)} & \rotatebox{90}{CIFAR-100 \cite{Krizhevsky:2009:LML}} & \rotatebox{90}{Caltech-101 \cite{Fei-Fei:2006:OSL}} & \rotatebox{90}{DTD \cite{Cimpoi:2014:DTW}} & \rotatebox{90}{Flowers \cite{Nilsback:2008:AFC}} & \rotatebox{90}{Pets \cite{Parkhi:2012:CAD}} & \rotatebox{90}{SVHN \cite{Netzer:2011:RDN}}  & \rotatebox{90}{Sun397 \cite{Xiao:2010:SDL}} & \rotatebox{90}{Camelyon \cite{Veeling:2018:REC}} & \rotatebox{90}{EuroSAT \cite{Helber:2019:END}}   & \rotatebox{90}{RESISC45 \cite{Cheng:2017:RSI}}  & \rotatebox{90}{Retinopathy \cite{Dugas:2015:DRD}} & \rotatebox{90}{CLEVR-Count \cite{Johnson:2017:CDD}} & \rotatebox{90}{CLEVR-Dist. \cite{Johnson:2017:CDD}}  & \rotatebox{90}{DMLab \cite{Beattie:2016:DML}} & \rotatebox{90}{KITTI-Dist. \cite{Geiger:2013:VMR}}  & \rotatebox{90}{dSpr-Loc. \cite{Matthey:2017:DDT}} & \rotatebox{90}{dSpr-Ori. \cite{Matthey:2017:DDT}}   & \rotatebox{90}{sNORB-Azi. \cite{LeCun:2004:LMG}}  & \rotatebox{90}{sNORB-Ele. \cite{LeCun:2004:LMG}} \\
\midrule
$r\!\in\![1..4]$ & 0.11 & 1 & 4 & 2 & 1 & 4 & 4 & 1 & 4 & 2 & 4 & 2 &4 & 2 & 4 & 4 & 4 & 4 & 4 & 4 \\
$r\!\in\![1..8]$ & 0.16 & 1 & 4 & 2 & 1 & 8 & 8 & 1 & 8 & 2 & 8 & 8 &4 & 8 & 8 & 8 & 8 & 4 & 8 & 8 \\
$r\!\in\![1..32]$ & 0.27 & 1 & 4 & 2 & 1 & 8 & 16 & 1 & 16 & 2 & 32 & 32 &4 & 8 & 8 & 8 & 32 & 4 & 32 & 8 \\
\bottomrule
\end{tabularx}
\endgroup

\end{table}

\begin{table}[t]
    \tabstyle{1pt}  %
    \caption{\textbf{Adapter rank $r$ for each FGVC dataset} for optimized versions of \adapter with different ranges of permitted ranks.}
    \label{tbl:adapter_ranks_fgvc}
    \vspace{-0.3em}
    \footnotesize
    \begingroup
\begin{tabularx}{\columnwidth}{@{}Xc@{\quad}@{\quad}c@{\quad}c@{\quad}c@{\quad}c@{\quad}c@{}}
\toprule
    & \rotatebox{90}{\#\,Param (M)} & \rotatebox{90}{CUB-200 \cite{Wah:2011:CBD}} & \rotatebox{90}{NABirds \cite{Horn:2015:BBR}} & \rotatebox{90}{Oxford Flowers \cite{Nilsback:2008:AFC}} & \rotatebox{90}{Stanford Dogs \cite{Khosla:2011:NDF} }& \rotatebox{90}{Stanford Cars \cite{Gebru:2017:FGC}} \\
\midrule
$r\!\in\![1..32]$ & 0.34 & 2 & 2 & 1 & 1 & 32 \\
\bottomrule
\end{tabularx}
\endgroup

    \vspace{-1em}
\end{table}

\section{Vision transformer pre-training}
As we add only very few parameters to an otherwise frozen backbone, the generalization capability of the feature representations produced by the backbone is important.
For ViTs, there are a number of off-the-shelf models available with differences in their training procedures.
Here, we examine three different pre-trainings as examples:
\begin{enumerate*}[label=\emph{(\arabic*)}]
    \item \emph{Original}: The ViT-B/16 weights used in the main paper, pre-trained with supervision on ImageNet-21k \cite{Russakovsky:2015:ILS} following the training procedure of the original ViT publication \cite{Dosovitskiy:2021:IWW},\footnote{\url{https://storage.googleapis.com/vit_models/imagenet21k/ViT-B_16.npz}}
    \item \emph{ImageNet-1k}: the same ViT weights further fine-tuned on ImageNet-1k \cite{Russakovsky:2015:ILS},\footnote{\url{https://storage.googleapis.com/vit_models/imagenet21k+imagenet2012/ViT-B_16-224.npz}} and
    \item \emph{AugReg}: weights from a pre-training with stronger data augmentation in the form of Mixup \cite{Zhang:2018:MBE} and Rand\-Aug\-ment \cite{Cubuk:2020:RPA} following \cite{Steiner:2022:HTY}.\footnote{\url{https://storage.googleapis.com/vit_models/augreg/B_16-i21k-300ep-lr_0.001-aug_medium1-wd_0.1-do_0.0-sd_0.0.npz}}
\end{enumerate*}

In \cref{tbl:pretraining}, we summarize our results for \adapter with rank $r\!=\!8$ evaluated on the VTAB validation sets.
We notice that additional fine-tuning on ImageNet-1k gives a slight edge (83.4\% average accuracy over 83.0\% for second best) in adaption for tasks that contain natural images. 
However, the fine-tuning is detrimental for the Specialized and Structured group.
Not fine-tuning on ImageNet-1k is beneficial for the Structured group with a large increase of \SI{3.7}{\pp}.
The Aug\-Reg training setting improves the transfer to the Specialized group but is worse than the other settings for natural images.
Overall, the original supervised training on ImageNet-21k generalizes best across all tasks in VTAB with an average accuracy of 76.5\%, \SI{0.3}{\pp} better than AugReg training and \SI{1.2}{\pp} better than ImageNet-1k fine-tuning.

\begin{table}[ht]
    \tabstyle{2pt}  %
    \caption{\textbf{Influence of ViT pre-training.} We use \adapter with rank $r\!=\!8$ for the evaluation and report the average accuracy in \% for each subgroup and across all groups on the VTAB \emph{val sets}.}
    \label{tbl:pretraining}
    \vspace{-0.3em}
    \footnotesize
    \begin{tabularx}{\columnwidth}{@{}Xc@{\enspace}c@{\enspace}c@{\qquad}c@{}}
\toprule
Pre-training & Natural & Specialized & Structured & Average \\
\midrule
ImageNet-1k & \textbf{83.4} & 86.5 & \underline{56.0} & 75.3 \\
AugReg & 81.6 & \textbf{87.2} & \textbf{59.7} & \underline{76.2} \\
Original & \underline{83.0} & \underline{86.8} & \textbf{59.7} & \textbf{76.5} \\
\bottomrule
\end{tabularx}

\end{table}

\section{Generality of the conclusions}
Using DINO \cite{Caron:2021:EPS} as an example of a ViT trained with self-supervision, we show in \cref{tbl:adapter_position_dino} that the orders of best-to-worst adapter position is consistent with that of a supervised backbone in terms of average accuracy, albeit with a higher standard deviation.
The ranking also stays the same for the comparison of \adapter with adapter configurations from previous work as presented in \cref{tbl:adapter_configurations_dino}.
This shows that our conclusions generalize beyond backbones with supervised pre-training to backbones based on self-supervised pre-training.

\begin{table}[t]
    \tabstyle{4pt}  %
    \caption{\textbf{Adapter position with DINO backbone.} We report average accuracy in \% ($\pm$~std.\ dev.) on the VTAB \emph{val sets} for different adapter positions. Adapter$_\text{base}$ with Houlsby initialization and rank $r\!=\!8$ is used in all experiments.}
    \label{tbl:adapter_position_dino}
    \vspace{-0.3em}
    \footnotesize
    \begin{tabularx}{\columnwidth}{@{}Xccc@{\qquad}c@{}}
\toprule
Position & Natural & Specialized & Structured & Average \\
\midrule
Pre & \underline{76.8} $\pm$ 0.4 & 86.2 $\pm$ 0.6 & 53.6 $\pm$ 0.7 & 72.2 $\pm$ 0.3 \\
Intermediate & \underline{76.8} $\pm$ 0.4 & 85.8 $\pm$ 0.8 & 52.6 $\pm$ 0.9 & 71.8 $\pm$ 0.4 \\
Parallel & 76.7 $\pm$ 0.3 & \textbf{86.8} $\pm$ 0.4 & \underline{54.1} $\pm$ 0.7 & \underline{72.5} $\pm$ 0.3 \\
Post & \textbf{76.9} $\pm$ 0.2 & \underline{86.3} $\pm$ 0.5 & \textbf{55.3} $\pm$ 0.7 & \textbf{72.8} $\pm$ 0.3 \\
\bottomrule
\end{tabularx}

\end{table}
\begin{table}[t]
    \tabstyle{4pt}  %
    \caption{\textbf{Comparison of \adapter with adapter configurations from previous work with DINO backbone.} We report the average accuracy in \% ($\pm$~std.\ dev.) of each subgroup and across all groups on the VTAB \emph{val sets}.}
    \label{tbl:adapter_configurations_dino}
    \vspace{-0.3em}
    \scriptsize
    \begin{tabularx}{\columnwidth}{@{}Xcc@{\;\:}c@{\;\:}c@{\enspace\quad}c@{}}
\toprule
Configuration & \#\,Param & Natural & Specialized & Structured & Average \\
\midrule
Houlsby \cite{Houlsby:2019:PET}, $r\!=\!8$ & 0.39 & \textbf{77.4} $\pm$ 0.4 & \textbf{86.5} $\pm$ 0.7 & 52.9 $\pm$ 0.8 & 72.3 $\pm$ 0.4 \\
Houlsby \cite{Houlsby:2019:PET}, $r\!=\!4$ & 0.24 & \underline{77.2} $\pm$ 0.5 & 86.2 $\pm$ 0.5 & 53.2 $\pm$ 0.8 & 72.2 $\pm$ 0.3 \\
Pfeiffer \cite{Pfeiffer:2021:AND}     & 0.21 & 76.8 $\pm$ 0.4 & 86.2 $\pm$ 0.3 & \underline{54.4} $\pm$ 1.0 & \underline{72.5} $\pm$ 0.4 \\
AdaptFormer \cite{Chen:2022:AAV}        & \textbf{0.19} & 76.5 $\pm$ 0.4 & 85.8 $\pm$ 0.4 & 53.0 $\pm$ 0.5 & 71.8 $\pm$ 0.3 \\
\adapter                            & \underline{0.20} & 76.7 $\pm$ 0.3 & \underline{86.4} $\pm$ 0.5 & \textbf{55.4} $\pm$ 0.8 & \textbf{72.8} $\pm$ 0.3 \\
\bottomrule
\end{tabularx}

\end{table}

\FloatBarrier

\end{document}